\documentclass{article}

\usepackage{arxiv}

\usepackage[utf8]{inputenc} 
\usepackage[T1]{fontenc}    
\usepackage{hyperref}       
\usepackage{url}            
\usepackage{booktabs}       
\usepackage{amsfonts}       
\usepackage{nicefrac}       
\usepackage{microtype}      
\usepackage{lipsum}
\usepackage{graphicx}
\usepackage{placeins}
\usepackage{amsmath}
\usepackage{subcaption}%
\graphicspath{ {./images/} }

\title{A light-weight model to generate NDWI from Sentinel-1}

\author{
  Saleh Sakib Ahmed \\
  Department of CSE, BUET\\
  Dhaka, Bangladesh \\
  \texttt{birdhunterx91@gmail.com} \\
   \And
  Saifur Rahman Jony \\
  Department of CSE, BUET\\
  Dhaka, Bangladesh \\
  \texttt{srj.buet17@gmail.com} \\
  \And
  Md. Toufikuzzaman \\
  Department of CSE, BUET\\
  Dhaka, Bangladesh \\
  \texttt{md.toufikzaman@gmail.com} \\
  \And
  Saifullah Sayed \\
  IWFM, BUET\\
  Dhaka, Bangladesh \\
  \texttt{saif.sayeed.11@gmail.com} \\
  \And
  Rashed Uz Zzaman \\
  IWFM, BUET\\
  Dhaka, Bangladesh \\
  \texttt{shoourov011@gmail.com} \\
  \And
  Sara Nowreen \\
  IWFM, BUET\\
  Dhaka, Bangladesh \\
  \texttt{snowreen@iwfm.buet.ac.bd} \\
  \And
  M. Sohel Rahman \\
  Department of CSE, BUET\\
  Dhaka, Bangladesh \\
  \texttt{msrahman@cse.buet.ac.bd} \\
}

\begin{document}
\maketitle
\begin{abstract}
 The use of Sentinel - 2 images to compute Normalized Difference water Index (NDWI) has many applications including water body area detection. However, cloud cover poses significant challenges in this regard, which hampers the effectiveness of Sentinel-2 images in this context. In this paper, we present a deep learning model that can generate NDWI given Sentinel - 1 images, thereby overcoming this ‘cloud barrier’.  We show the effectiveness of our model, where it demonstrates a high accuracy of 0.9134 and an AUC of 0.8656 to predict the NDWI. Additionally, we observe promising results with an R2 score of 0.4984 (for regressing the NDWI values) and a Mean IoU of 0.4139 (for the underlying segmentation task). In conclusion, our model offers a first and robust solution for generating NDWI images directly from Sentinel - 1 images and subsequently use for various application even under challenging conditions such as cloud cover and nighttime.
\end{abstract}


\section{Introduction}
In 1996, Gao et al. published the foundational paper introducing the concept of the Normalized Difference Water Index (NDWI) for remote sensing of vegetation liquid water from space \cite{gao1996ndwi}. This innovative approach utilized multispectral satellite images to calculate the water index, as shown in Eq. \ref{eq:NDWI}. Subsequently, McFeeters proposed a variant of NDWI specifically focusing on water bodies, highlighting its utility in enhancing open water features in satellite imagery \cite{mcfeeters1996use}. In 2010, Chandrasekar et al. explored the relationship between NDWI and rainfall, providing valuable insights into its application for drought monitoring \cite{chandrasekar2010land}. These developments have solidified NDWI as a highly useful tool for various environmental and hydrological applications.

However, significant challenges persist in this domain. The signals required to compute NDWI are highly susceptible to interference from cloud cover and nighttime conditions, which can severely limit its reliability. This issue particularly affects Sentinel-2, a satellite commonly used to calculate NDWI. In contrast, Sentinel-1, which operates using radar signals, is immune to such limitations, making it a promising alternative source of data.

In this research, we aim to overcome these barriers by developing and presenting a lightweight machine learning model based on the U-Net architecture \cite{ronneberger2015u}. This model is designed to convert Sentinel-1 images into NDWI, enabling the generation of NDWI even when Sentinel-2 images are unavailable or compromised due to clouds or other obstructions. By leveraging the complementary strengths of Sentinel-1 and machine learning, we provide a robust solution for generating NDWI under challenging conditions, thereby expanding its applicability to a broader range of scenarios.

\section{Methods}

\subsection{Problem Description}

As mentioned previously, we aim to compute the NDWI even in overcast environments. To achieve this, we utilize Sentinel-1 images, which are immune to cloud cover due to their use of radar signals. Unlike Sentinel-2, which captures multispectral optical images and is susceptible to interference from clouds, Sentinel-1 is a radar imaging satellite that operates in the C-band, providing consistent data regardless of weather or lighting conditions. Sentinel-1 data consists of two channels: VV (vertically transmitted and vertically received) and VH (vertically transmitted and horizontally received), which provide detailed information about surface properties.

Using these two-channel radar images from Sentinel-1, we aim to generate NDWI. For simplicity, we have streamlined the task by employing Otsu’s thresholding method \cite{otsu1975threshold} (described in detail below), which maximizes inter-class variance in cloud-free NDWI images. This preprocessing step simplifies the classification of water and non-water regions, ensuring robustness in the generated NDWI. 

\begin{figure*}[t]
\begin{center}
        \includegraphics[width=0.8\textwidth]{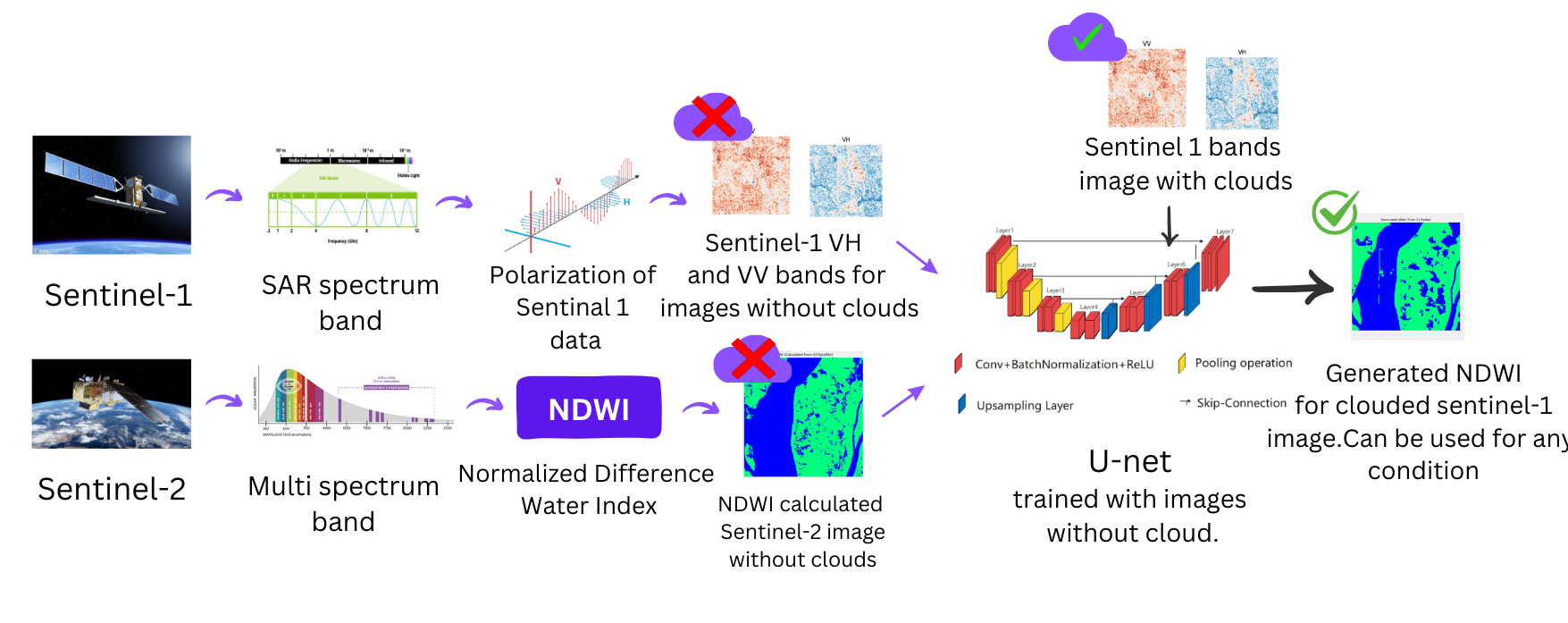}

\end{center}
   \caption{ The Figure showcases the procedure of how we train our CloudBreaker model. We use Sentinel-1 images as input and the NDWI of corresponding Sentinel-2 images as ground truth for the U-Net model (only cloud-free images are considered). After training, CloudBreaker becomes capable of directly producing NDWI images of corresponding Sentinel-2 images from given Sentinel-1 images}
    \label{fig:methods}
\end{figure*}

Ultimately, our model enables the production of NDWI images from Sentinel-1 data under any conditions, overcoming the limitations imposed by cloud cover. The overall methodology is visualized in Figure \ref{fig:methods}.

\subsection{Datasets}

In this research, the Cloud to Street - Microsoft Flood and Clouds Dataset made publicly available by Radiant Earth Foundation \cite{cloud2024} has been utilized. This dataset comprises 900 pairs of chips from Sentinel-1 and Sentinel-2, gathered across 18 international flooding incidents. The Sentinel-1 chips feature two spectral bands, VV and VH, whereas the Sentinel-2 chips include 13 spectral bands including the Red, Green, Blue, and Near-Infrared (NIR) bands. The chips are 512 x 512 pixels each in size and capture images under both clear and cloudy conditions. Additionally, the dataset provides masks for water and clouds for each chip.

\subsection{Data Preprocessing}

The original $512\times512$ pixel chips from the dataset were segmented into 16 smaller $128\times128$ pixel chips, expanding the total count of Sentinel-1 and Sentinel-2 chip pairs from 900 to 14,400. This augmentation significantly increases the volume of data available, making it suitable for training a deep neural network with a more compact architecture. Cloud masks were applied to eliminate any chip that displayed cloud cover, resulting in a final tally of 7,878 chips for the purpose of training and validation. The data for each chip was normalized to a scale between 0 and 1. For the development of our model, 80\% of this cloud-filtered dataset was allocated for training (training set), with the remaining 20\% used for test purposes (test set).

The Sentinel-1 data consists of two polarization bands: VV (vertically transmitted and vertically received) and VH (vertically transmitted and horizontally received), each represented as a 2D array with a resolution of \(128 \times 128\) (width \(\times\) height). These bands are combined along the channel dimension, forming a 3D input array of shape \((128, 128, 2)\), where:
\begin{itemize}
    \item \(128\): Image width (pixels).
    \item \(128\): Image height (pixels).
    \item \(2\): Channels corresponding to VV and VH.
\end{itemize}

For batch processing, the inputs are arranged into a 4D array with shape \((\text{batch\_size}, 128, 128, 2)\), where \(\text{batch\_size}\) is the number of images in the batch. For example, a batch size of \(32\) results in an input shape of \((32, 128, 128, 2)\). To compute the NDWI, we utilized the Green and NIR bands from the Sentinel-2 chips, applying the following equation:
\begin{equation}
\label{eq:NDWI}
\text{NDWI} = \frac{\text{Green} - \text{NIR}}{\text{Green} + \text{NIR}}
\end{equation}
The NDWI values, which range between -1 and 1, were rescaled to a 0 to 1 range.

\subsection{Otsu Thresholding}
Otsu's method finds the optimal threshold to split an image into foreground/background by maximizing inter-class variance. Key steps:
\begin{itemize}
    \item Convert histogram to probabilities: $p(i) = \frac{n(i)}{N}$ where $n(i)$ = pixels at level $i$, $N$ = total pixels.
    \item Compute cumulative probabilities $\omega(t) = \sum_{i=0}^{t} p(i)$ and cumulative means $\mu(t) = \sum_{i=0}^{t} i \cdot p(i)$.
    \item Total mean: $\mu_T = \sum_{i=0}^{L-1} i \cdot p(i)$
    \item Optimal threshold $t^*$ maximizes inter-class variance:
    \[
    t^* = \arg\max_t \frac{(\mu_T \omega(t) - \mu(t))^2}{\omega(t)(1-\omega(t))}
    \]
\end{itemize}
Maximizing variance separates distinct pixel groups effectively. This is used to simplify the task.

\subsection{Model Architecture}

The U-Net \cite{ronneberger2015u} processes \textbf{128×128×2 inputs} through an encoder-decoder structure: \\
\textbf{Encoder} (4 downsampling blocks: 64→128→256→512 filters) with max-pooling → \\
\textbf{Bottleneck} (1024 filters) → \\
\textbf{Decoder} (4 upsampling blocks: 512→256→128→64 filters) using transposed convolutions and skip connections. \\
Final output is generated through a \textbf{3×3 convolution} (He-initialized) with \textbf{sigmoid activation}, producing \textbf{128×128×1} segmentation masks. Symmetric architecture preserves spatial details via encoder-decoder skip connections.

\subsection{Evaluation Metrics}
The chosen metrics address distinct aspects of our NDWI generation post otsu thresholding pipeline: \textbf{Loss} optimizes NDWI value prediction accuracy; \textbf{R\textsuperscript{2}} validates SAR-to-NDWI regression quality; \textbf{AUC} assesses class separability critical for Otsu's thresholding; \textbf{Accuracy} measures post-threshold classification despite class imbalance; and \textbf{Mean IoU} directly evaluates water boundary segmentation precision. Together they verify both the regression fidelity (essential for physical interpretation of Sentinel-1 radar backscatter) and segmentation utility (for practical water mapping), while AUC specifically ensures robust threshold detection in imbalanced water/non-water distributions.

\subsection{Code, Environment, and Availability}

We used Colab Pro with a T4 GPU, 51 GB of system RAM, and 15 GB of GPU RAM to run our experiments. The duration for each epoch was roughly 13 seconds. All our Code can be found in the following link: \url{https://github.com/bojack-horseman91/CloudBreaker}

\section{Results and Discussions}


As shown in Table~\ref{table:performance} tracking metric evolution during training, our model demonstrates strong generalization for NDWI estimation from SAR data. The high test accuracy (0.9134) and AUC (0.8656) indicate robust pixel-wise classification of water bodies post-Otsu thresholding. While training-test R\textsuperscript{2} disparity (0.7602 vs. 0.4984) suggests some regression overfitting, the test score remains reasonable for SAR-to-NDWI mapping. Model selection based on validation loss prioritized generalization over peak R\textsuperscript{2} scores observed in intermediate epochs. Consistent IoU values (0.4310 train vs. 0.4139 test) confirm stable segmentation performance across datasets, despite greater test set variability in regression metrics.

\begin{figure}[t]
\begin{center}
        \includegraphics[width=0.5\textwidth]{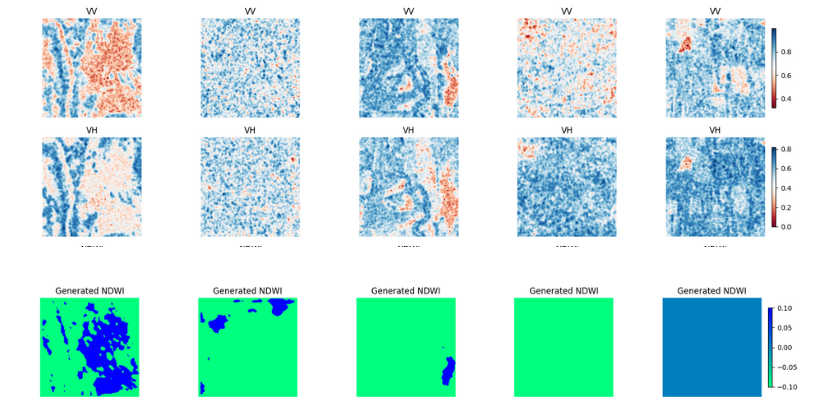}

\end{center}
   \caption{Model was trained using cloud-free Sentinel-1 data and Sentinel-2 images. Finally, we tested it with Sentinel-1 data containing clouds to demonstrate its ability to generate NDWI images of Sentinel-2 when such images are not properly available. The generated NDWI, as well as the two bands of Sentinel-1 data, VV and VH, are shown.}

    \label{fig:cloud_images}
\end{figure}

\begin{figure}[h!]
    \centering
    \begin{subfigure}[b]{0.5\textwidth}
        \centering
        \includegraphics[width=\textwidth]{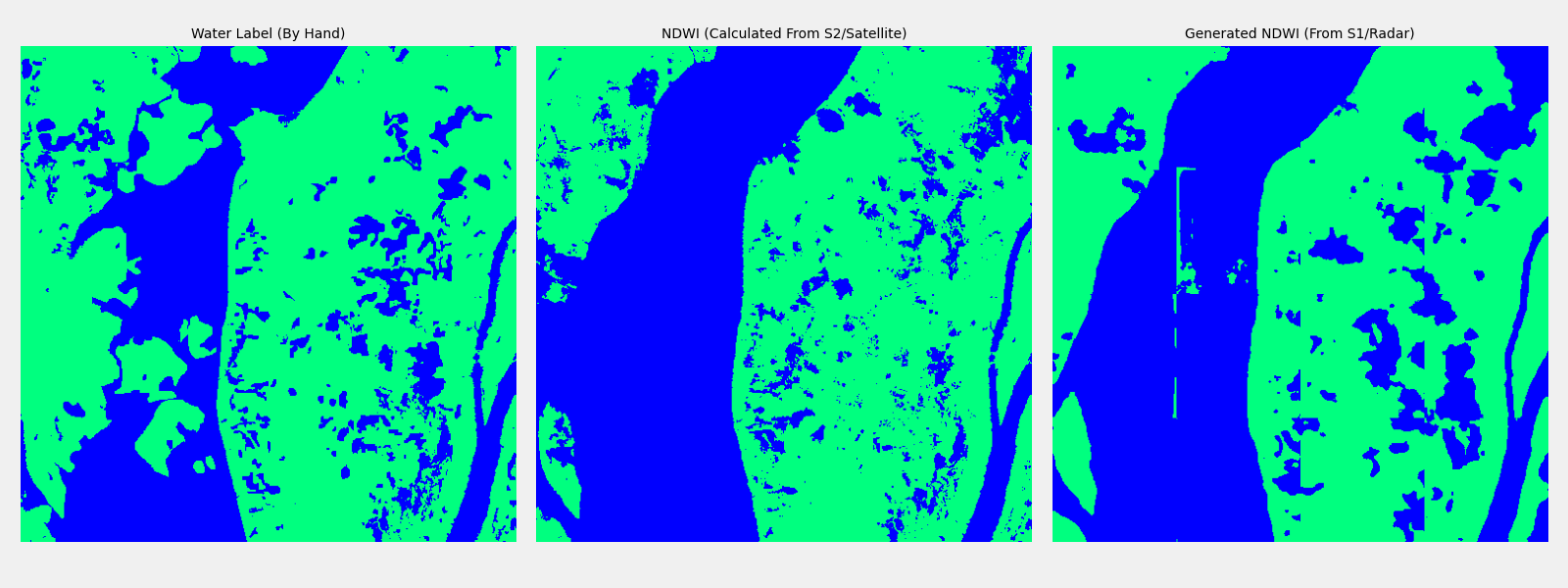}
        \caption{}
        \label{fig:gen_a}
    \end{subfigure}
    \begin{subfigure}[b]{0.5\textwidth}
        \centering
        \includegraphics[width=\textwidth]{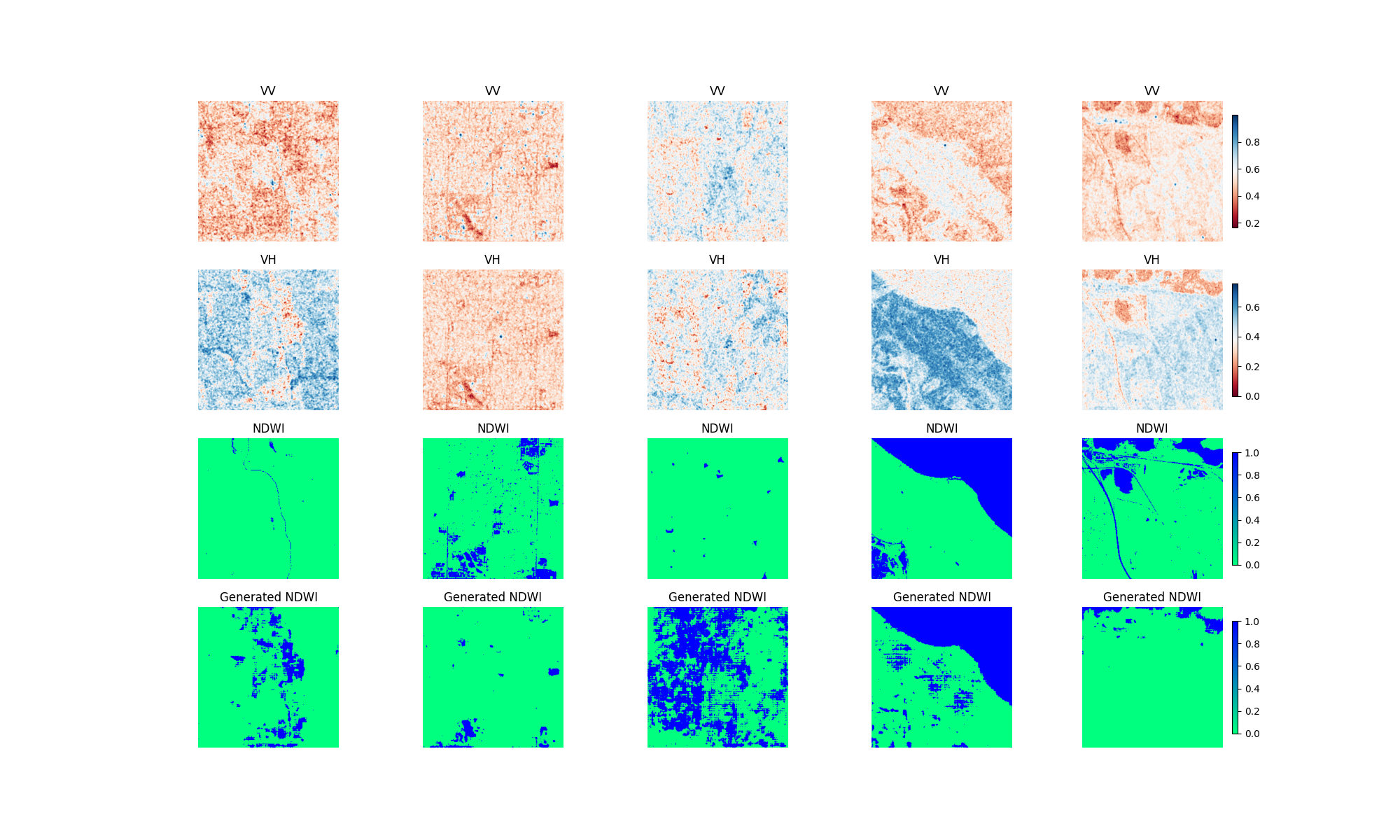}
        \caption{}
        \label{fig:gen_b}
    \end{subfigure}
    
    \caption{(\textbf{a}) The model was trained with Sentinel-1 images and ground truth as Sentinel-2 images. The generated image is on the rightmost, with the Sentinel-2 image in the middle and the hand-labeled image on the leftmost. (\textbf{b}) This image shows how the two bands of Sentinel-1 images, VH and VV, are used to train the model, along with the NDWI of Sentinel-2 images, to generate NDWI for those Sentinel-1 images.}

    \label{fig:gen}
\end{figure}

Visual comparisons in Figures~\ref{fig:gen_a}--\ref{fig:gen_b} demonstrate substantial visual congruence between generated NDWI from Sentinel-1 (VV/VH bands) and reference Sentinel-2 products. Crucially, Figure~\ref{fig:cloud_images} reveals our model's operational advantage: reliable NDWI generation under cloud cover using SAR data where optical methods fail. Cloud cover Sentinel-2 would not be able to get the correct NDWI.To our knowledge, this constitutes the first demonstrated end-to-end learning framework for direct NDWI estimation from Sentinel-1 imagery, circumventing optical limitations through SAR-physics-informed deep learning. The technical approach enables continuous water monitoring regardless of atmospheric conditions - a critical advancement over existing optical-based approaches.
\begin{table}[h!]
\centering
\begin{tabular}{|c|c|c|c|c|}
\hline
\textbf{Metric} & \textbf{Accuracy} & \textbf{AUC} & \textbf{R\textsuperscript{2} Score} & \textbf{Mean IoU} \\ \hline
\textbf{Training} & 0.9575 & 0.9755 & 0.7602 & 0.4310 \\ \hline
\textbf{Testing}  & 0.9134 & 0.8656 & 0.4984 & 0.4139 \\ \hline
\end{tabular}
\caption{Performance of our model in different metrics for training and test data}
\label{table:performance}
\end{table}
\section{Conclusions}

Our model is able to generate NDWI images from Sentinel-1 images with high efficacy for both normal as well as cloudy conditions. So it promises to break the barrier of clouds (and or other related conditions) faced by temporal models that are dependent on Sentinel-2 images. This study and hence our model is the first of its kind and promises to spark subsequent works where it will be used as the baseline for further benchmarking. Along that line, as an immediate future work, we plan to extend our benchmarking using other models in parallel to U-Net, particularly using generative architecture.

\FloatBarrier
\bibliographystyle{plain} 

\bibliography{references}

\begin{thebibliography}{1}

\bibitem{chandrasekar2010land}
K~Chandrasekar, MVR Sesha~Sai, PS~Roy, and RS~Dwevedi.
\newblock Land surface water index (lswi) response to rainfall and ndvi using the modis vegetation index product.
\newblock {\em International Journal of Remote Sensing}, 31(15):3987--4005, 2010.

\bibitem{cloud2024}
{Cloud to Street - Microsoft Flood and Clouds Dataset}.
\newblock {Cloud to Street - Microsoft Flood and Clouds Dataset}.
\newblock \url{https://registry.opendata.aws/c2smsfloods}.
\newblock Accessed: Feb 14, 2024.

\bibitem{gao1996ndwi}
Bo-Cai Gao.
\newblock Ndwi—a normalized difference water index for remote sensing of vegetation liquid water from space.
\newblock {\em Remote sensing of environment}, 58(3):257--266, 1996.

\bibitem{mcfeeters1996use}
Stuart~K McFeeters.
\newblock The use of the normalized difference water index (ndwi) in the delineation of open water features.
\newblock {\em International journal of remote sensing}, 17(7):1425--1432, 1996.

\bibitem{otsu1975threshold}
Nobuyuki Otsu et~al.
\newblock A threshold selection method from gray-level histograms.
\newblock {\em Automatica}, 11(285-296):23--27, 1975.

\bibitem{ronneberger2015u}
Olaf Ronneberger, Philipp Fischer, and Thomas Brox.
\newblock U-net: Convolutional networks for biomedical image segmentation.
\newblock In {\em Medical image computing and computer-assisted intervention--MICCAI 2015: 18th international conference, Munich, Germany, October 5-9, 2015, proceedings, part III 18}, pages 234--241. Springer, 2015.

\end{thebibliography}

\end{document}